\documentclass[letterpaper, 10 pt, conference]{ieeeconf}  

\IEEEoverridecommandlockouts                              

\usepackage{graphics} 
\usepackage{epsfig} 
\usepackage{mathptmx} 
\usepackage{times} 
\usepackage{amsmath} 
\usepackage{amssymb}  
\usepackage[dvipsnames]{xcolor}
\usepackage[space]{cite}    
\usepackage{tabularx}
\usepackage{algorithm}
\usepackage{algpseudocode}
\usepackage{multirow}
\usepackage{caption}
\usepackage[hidelinks]{hyperref}
\usepackage[capitalize]{cleveref}  

\usepackage{eso-pic}

\AddToShipoutPictureBG{%
  \AtPageUpperLeft{%
    \put(\LenToUnit{0.25\paperwidth+0.375in},\LenToUnit{-12mm}){%
      \parbox{\textwidth}{\fontsize{10}{12}\selectfont PREPRINT OF PAPER ACCEPTED TO IEEE ICRA 2024}%
    }%
  }%
}

\begin{document}

\title{\LARGE \bf
Tactile-Informed Action Primitives Mitigate Jamming in Dense Clutter
}

\author{Dane Brouwer, Joshua Citron, Hojung Choi, Marion Lepert,
Michael Lin, Jeannette Bohg, and Mark Cutkosky
\thanks{The authors are with Stanford University, USA
        {\tt\footnotesize \{daneb, jcitron, hjchoi92, letpertm, mlinyang, bohg, cutkosky\} @stanford.edu}}%
}

\maketitle
\thispagestyle{empty}
\pagestyle{empty}

\begin{abstract}
It is difficult for robots to retrieve objects in densely cluttered lateral access scenes with movable objects as jamming against adjacent objects and walls can inhibit progress. We propose the use of two action primitives---burrowing and excavating---that can fluidize the scene to un-jam obstacles and enable continued progress. Even when these primitives are implemented in an open loop manner at clock-driven intervals, we observe a decrease in the final distance to the target location. Furthermore, we combine the primitives into a closed loop hybrid control strategy using tactile and proprioceptive information to leverage the advantages of both primitives without being overly disruptive. In doing so, we achieve a 10-fold increase in success rate above the baseline control strategy and significantly improve completion times as compared to the primitives alone or a naive combination of them.

\end{abstract}
\section{Introduction}

Conventional robot navigation and path planning seek to avoid obstacles \cite{pandey2017mobile}, which limits the ability to interact with densely cluttered environments. As a result, navigation among movable obstacles (NAMO) has become an active area of research \cite{stilman2005navigation, stilman2007manipulation, wang2020affordance, ellis2022navigation}. Even so, the scenes that robots are presented with in the 
literature rarely contain substantial clutter. Robots must interact with dense clutter when operating among the shelves and cupboards of our homes, in the rubble of collapsed buildings during search and rescue, or in natural environments for sample collection. 

Navigating clutter made up of movable objects appears in the literature primarily in the form of 
table-top manipulation \cite{srivastava2014combined,muhayyuddin2017randomized,huang2021visual} and lateral access reaching \cite{gupta2013interactive,huang2021mechanical,huang2022mechanical}, in part due to the prevalence of such scenes in factories, warehouses, and homes. 
Object retrieval in these scenes can be difficult, with task requirements ranging from identifying or predicting target object locations to reaching past obstacles toward desired locations and retracting once objects are acquired. 

Approaches for establishing target locations generally rely on vision and are therefore greatly affected by occlusions when scenes are cluttered and constrained. This is common for lateral access tasks, in which the scene is only accessible from one side. Visual information is especially limited while a manipulator is interacting with the scene, partially blocking its own view. Consequently, simultaneous action and observation are often infeasible. In this case, a typical approach is to perform mechanical search by iteratively observing, performing open loop actions, and retracting to re-gather state information of the scene \cite{huang2021mechanical,huang2022mechanical}. 

Iterative mechanical search is time intensive, hence it is desirable to develop methods that can accurately reach target locations based on an initial observation of the scene, eliminating the need for multiple search cycles. In some cases, the scene is so densely cluttered that clearing a visual path to the target is impractical, or the target object has spectral properties that make it difficult to identify with vision. To tackle these cases, several works propose methods to identify target objects using tactile information \cite{schneider2009object, xu2013tactile, lin2019learning, zhong2022soft}. 

\begin{figure}[t!]
\centering
	\includegraphics[width=3.35in]
 {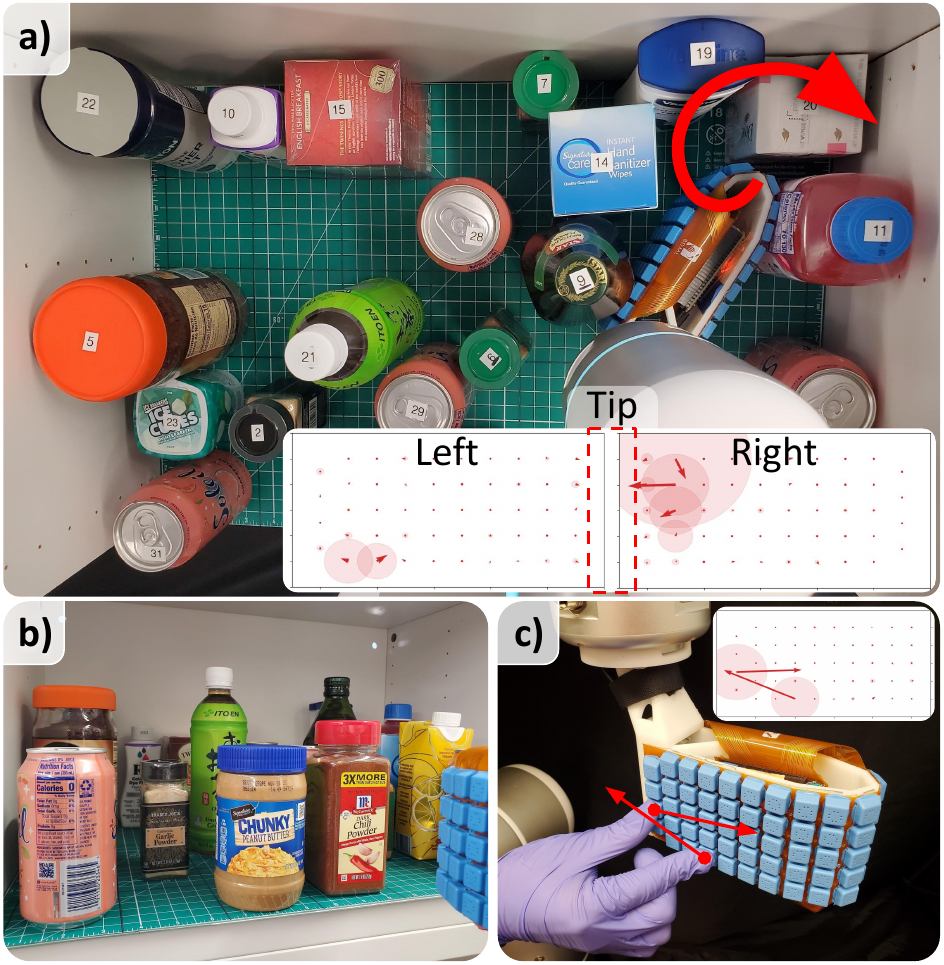}
	\caption{a) An end-effector, covered with soft tactile sensors, reaches toward a target in a cupboard with numerous objects. Using motion primitives can prevent object jamming. Inset images show contact forces---circle diameter represents normal force and arrows represent shear forces---on each side of the end-effector. In this case, the contact forces have triggered a \emph{clockwise excavate} primitive indicated by the red, curved arrow. b) Typical front view available to the robot for a lateral access scene. c) Image of the right side of the end-effector being pinched and corresponding inset force visualization.}
	\label{fig:Sensor}
    \label{fig:CoverPhoto}
	\vspace{-15pt}
\end{figure}

This paper focuses on enabling robots to perform a reaching maneuver toward expected target locations without the need for vision. This approach removes the issue of self-occlusion during reaching and can provide robustness to approaches that continue to leverage vision throughout the task. 
The scenario we study in this paper is shown in \cref{fig:CoverPhoto} with a finger-like appendage covered with tactile sensors reaching into a crowded cupboard.
Prior research has enabled vision-less reaching by equipping manipulators with tactile sensors and reaching to desired locations while modulating contact forces using model predictive control \cite{jain2013reaching, bhattacharjee2014robotic, killpack2016model, albini2021exploiting}. In this prior work, however, the scenes consist of objects with fixed bases, so scenes are static and scene-to-scene variation is limited. 
In contrast, we consider the task of reaching among movable objects in a fixed space (e.g. a cupboard), where the objects may sporadically resist motion as they jam against adjacent objects and walls. In these scenes, the objects are so densely packed that there is no collision free path toward the target location and a strategy that disturbs the object arrangement is unavoidable.

\paragraph*{Contributions}

In this paper, we present the use of two new action primitives that reduce the likelihood that robot motion will be impeded by object jamming. Additionally, we investigate how sensor feedback can inform when to deploy them. The primitives achieve similar performance in simulation and on hardware using a robot arm with an array of soft triaxial tactile sensors. An event-based approach based on contact forces and locations can help to avoid jamming in severe cases where running the primitives in an open-loop manner may fail. 
The results of this study can enable mobile robot NAMO and reaching through dense collections of movable objects using compact end-effectors.

\section{Methods}

We present two action primitives---burrow and excavate---to mitigate jamming when reaching for objects in a cluttered, lateral access scenario. We argue that deploying these primitives in a cohesive, tactile-informed manner provides an additional benefit over deploying the primitives in isolation. In the following sections we explain the details of these implementations.

We simplify the problem by taking a known target location as a prior and focusing on strategies that a single manipulator can employ. The scenario is analogous to creating a single ``virtual finger'' \cite{iberall1987nature} as humans often do when bringing their index through fifth fingers together to reach into a cupboard or other cluttered space to retrieve a hidden object.

\subsection{Straight Line Control}
\label{sec:straight}

A simple method to command motions of the robot is to send an angular velocity command that points the link toward the target while sending a velocity command in the direction of the target location. We implement straight line control by sending velocity commands according to
\begin{equation}
    \Vec{v}_{SL} = v_{max}\hat{d}_{targ}
    \label{eqn:SL_lin}
\end{equation}
where $\Vec{v}_{SL}$ is the commanded straight line linear velocity, $v_{max}$ is the maximum allowed velocity magnitude, and, as shown in \cref{fig:PrimSchem}a, $\hat{d}_{targ}$ is the unit vector pointing from the tip to the target. Angular velocity commands are implemented according to
\begin{equation}
    \omega_{SL} = sign(\phi - \theta) \omega_{max}
    \label{eqn:SL_ang}
\end{equation}
where $\omega_{SL}$ is the commanded straight line angular velocity, $\omega_{max}$ is the maximum allowable angular velocity, and, as shown in \cref{fig:PrimSchem}a, $\phi$ is the angle of $\hat{d}_{targ}$ w.r.t. the $y$-axis, and $\theta$ is the angle of the end-effector w.r.t. the $y$-axis. For each strategy, the maximum velocity magnitudes are decreased when approaching the goal to avoid unstable behavior. 

Straight line control attempts to progress directly toward the goal without deviation and therefore succeeds quickly in easy scenarios but will be susceptible to getting stuck due to object jamming. As such, it is used as the baseline to compare performance. 

\subsection{Burrow Primitive}
\label{sec:burrow}
\begin{figure}[ht]
\centering
	\includegraphics[width=3.35in]{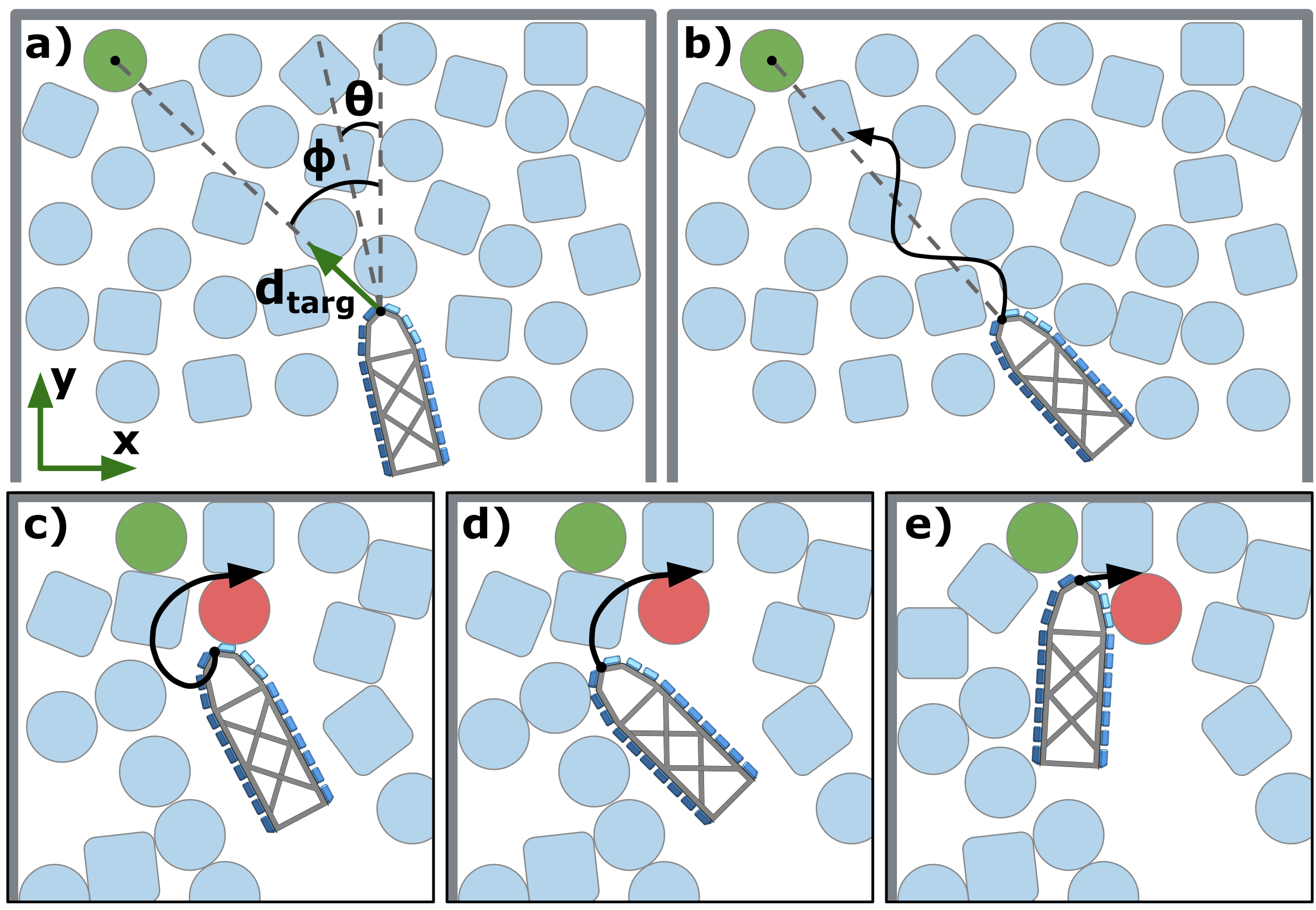}
	\caption{Schematic visualizations of a) straight line control, b) burrowing action primitive and c--e) a sequence showing the progression of a clockwise excavate action primitive.
}
 
	\label{fig:PrimSchem}
\end{figure}

The burrowing primitive adds sinusoidal lateral motions to the straight line command in order to generate a snaking behavior (\cref{fig:PrimSchem}b). The behavior is qualitatively similar to that used by some worms and other animals to reduce the effort of burrowing through granular media \cite{ortiz2019soft,das2023effects}.

This approach clears a corridor and perturbs impeding objects to reduce the likelihood of jamming. It also allows compliant contour following if an impeding object is only resisting progress on one side of the link. 
This behavior occurs naturally while burrowing since motion will be prevented in the impeding direction but free in the opposite lateral direction. As a result, the link will make and break contact with the impeding object and during the time that it is not in contact, progress toward the goal can continue.

The linear velocity commands for the burrowing primitive are implemented through
\begin{equation}
    \Vec{v}_{bur} = v_{max}\frac{\Vec{v}_{SL} + \Vec{v}_{sin}}{||\Vec{v}_{SL} + \Vec{v}_{sin}||}
    \label{eqn:BU_lin}
\end{equation}
where $\Vec{v}_{sin}$ is perpendicular to $\Vec{v}_{SL}$ and is scaled by a sinusoid function as follows 
\begin{equation}
    \Vec{v}_{sin} =
    \left( \hat{d}_{targ} \times \hat{z} \right)
    \left( \frac{A_{bur}}{1-A_{bur}} \right) \sin(f_{bur}t)
    \label{eqn:BU_lin2}
\end{equation}
with the burrow amplitude, $A_{bur}$, corresponding to straight line motion at $A_{bur} = 0$ and purely side-to-side motion as $A_{bur}$ approaches 1. $f_{bur}$ is the burrowing frequency and $t$ is the time since the reaching task began. The angular velocity $\omega_{bur}$ is commanded according to \cref{eqn:SL_ang}.

\subsection{Excavate Primitive}
\label{sec:excavate}
If objects become jammed despite burrowing, an excavate primitive may free them.
The excavate primitive is a scooping motion such that the link translates in a spiral (\cref{fig:PrimSchem}c) and rotates. If the spiral is counter-clockwise (CCW), the link rotates CCW, and vice-versa. In this way, the excavate maneuver tends to push impeding obstacles to the side (right if CW; left if CCW). This primitive is ideally performed either to clear a jammed object that impedes motion or prevent downstream jamming. 
This maneuver is performed for a set length of time. For a CCW excavate, while $t - t_{excv, start} \leq t_{excv}$, the commanded velocity goes as:
\begin{equation}
    \Vec{v}_{excv} = K v_{max} R_{\theta} 
    \begin{bmatrix}
        \sin(\frac{3\pi}{2}t_{frac}) - L\omega_{excv} \\
        -\cos(\frac{3\pi}{2}t_{frac}) \\
    \end{bmatrix}
    \label{eqn:EX_lin}
\end{equation}
with 
\begin{equation}
    K = \frac{1 + (s_{excv} - 1)t_{frac}}{s_{excv}} 
    \label{eqn:EX_lin2}
\end{equation}
where $R_{\theta}$ is the rotation matrix according to the end-effector's current orientation, $\theta$; $L$ is the length of the end-effector, $t_{frac} = (t - t_{excv, start})/t_{excv}$ is the fractional time during the spiral maneuver, and $s_{excv}$ is a scaling ratio used to adjust how rapidly the radius grows during the spiral motion. If $s_{excv} = 1$, the motion is a pure arc of $\frac{3\pi}{2}$\,radians. If $s_{excv} > 1$, a partial spiral is formed, where the commanded tip velocity grows from 
$v_{max}/s_{excv}$ at the beginning of the excavate maneuver to $v_{max}$ at the end. In this study, we choose $s_{excv} = 2$.

The angular velocity is likewise commanded according to a growing sinusoid to first get around an impeding object (\cref{fig:PrimSchem}d) and then push it to the side (\cref{fig:PrimSchem}e):
\begin{equation}
    \omega_{excv} = -K\omega_{max}\sin(2\pi t_{frac})
    \label{eqn:EX_ang}
\end{equation}
with $K$ from \cref{eqn:EX_lin2}.
For a CW excavate, the sign of the $x$-component of linear velocity and the sign of the angular velocity are flipped.

\subsection{Control Strategies}
\label{sec:strategies}

While the previous sections describe the motion primitives, this section explains the control strategies in this paper and how the primitive motions are combined to form them.

\subsubsection{Primitives in Isolation}
\label{subsec:OStrats}

We implement three control strategies that deploy the primitives in isolation. The \textbf{straight line} control strategy continuously commands motions according to \cref{eqn:SL_lin}-(\ref{eqn:SL_ang}). The \textbf{burrow} control strategy continuously performs motions as described in \cref{eqn:SL_ang}-(\ref{eqn:BU_lin2}). The \textbf{excavate} control strategy performs an excavate---as described in \cref{eqn:EX_lin}-(\ref{eqn:EX_ang})---at regular intervals, $t_{trig}$, and resumes straight line control otherwise. The direction of the excavation is randomly chosen each time. 

The key parameters that adjust the strategies' behavior are the burrowing amplitude, $A_{bur}$, burrowing frequency, $f_{bur}$, excavate duration, $t_{excv}$, and excavate trigger interval, $t_{trig}$.

\subsubsection{Hybrid Strategies}

We develop two hybrid control strategies to explore the benefits of using tactile feedback to decide when to execute primitives. 

The \textbf{hybrid clock} control strategy deploys the excavate primitive at fixed, clock-driven intervals as before (i.e. every $t_{trig}$ seconds) and executes the burrow control strategy otherwise.

The \textbf{hybrid event} control strategy triggers primitive motions according to event-driven, sensory conditions. This strategy performs an excavate in two sets of circumstances: (1) if a light, extended contact 
($F_{push, min}$ $\leq$ $F$ $\leq$ $F_{push, max}$ for $t_{push}$ or longer)
is detected at the tip of the end-effector or (2) if progress toward the goal slows substantially and a large contact force is detected (the distance to goal has not been reduced by $v_{max}(t - t_{prog})$ and $F$ $\geq$ $F_{excv}$).
Case (1) will usually occur if an object is being pushed to the back of the scene; in this case an excavate primitive can potentially forestall future jamming. Case (2) usually signals that progress has stopped due to object jamming. The location of the peak contact force (i.e., whether on the left or right side of the end-effector) affects whether the strategy performs a CCW or CW excavate procedure, respectively.

When not excavating, the hybrid event control strategy will burrow if a moderate force threshold, $F_{bur}$,
is exceeded but will resume the straight line control strategy otherwise. In this way, a direct path to the goal is taken when possible.

\section{Experiment Setup}
We perform experiments both on hardware and in simulation. This section provides details about both setups and specifies the experiments we conduct. 

We constrain the task such that the robot must complete its reaching within a predefined time limit, $t_{tot}$. We use velocity control with a maximum linear velocity magnitude, $v_{max}$, and angular velocity, $\omega_{max}$, and do not allow the robot to exceed a desired maximum in-plane wrench: [$F_x < F_{max}$, $F_y < F_{max}$, $M_z < M_{max}$].

\subsection{Hardware}

\subsubsection{Sensorized end-effector}
The control strategies as described in \cref{sec:strategies} are deployed on a sensorized, finger-like end-effector, as seen in \cref{fig:Sensor}c. The sensor provides triaxial contact force information from each taxel at 25\,Hz. It does this by reading magnetic flux changes caused by deformations of magnets embedded in elastomer structures. Fabrication, communication, and additional characterization details can be found in \cite{choi2022deep}.

Unlike the reported previous application, it is now necessary to compensate for changes in orientation with respect to the earth's magnetic field. To separate this effect, we periodically perform a compensation at $\approx 3$\,Hz as follows. We represent each force component of each sensor array (e.g. the $x$-component of Sensor Array 1) as a surface where each taxel's measured force corresponds to a height measurement of the surface at that taxel location. In contrast to highly localized variations due to contacts, we expect measurements that are consistent across the surface to correspond to bulk magnetic field changes. Therefore, we  fit a plane to these measurements using RANSAC then subtract it from the reading of each taxel. This process is performed for all three force components on each sensor array. The resulting data are communicated at $\approx$15\,Hz to the robot controller.

\subsubsection{Lateral Access Environment}

\begin{figure}[htb]
\centering
	\includegraphics[width=3.0in]{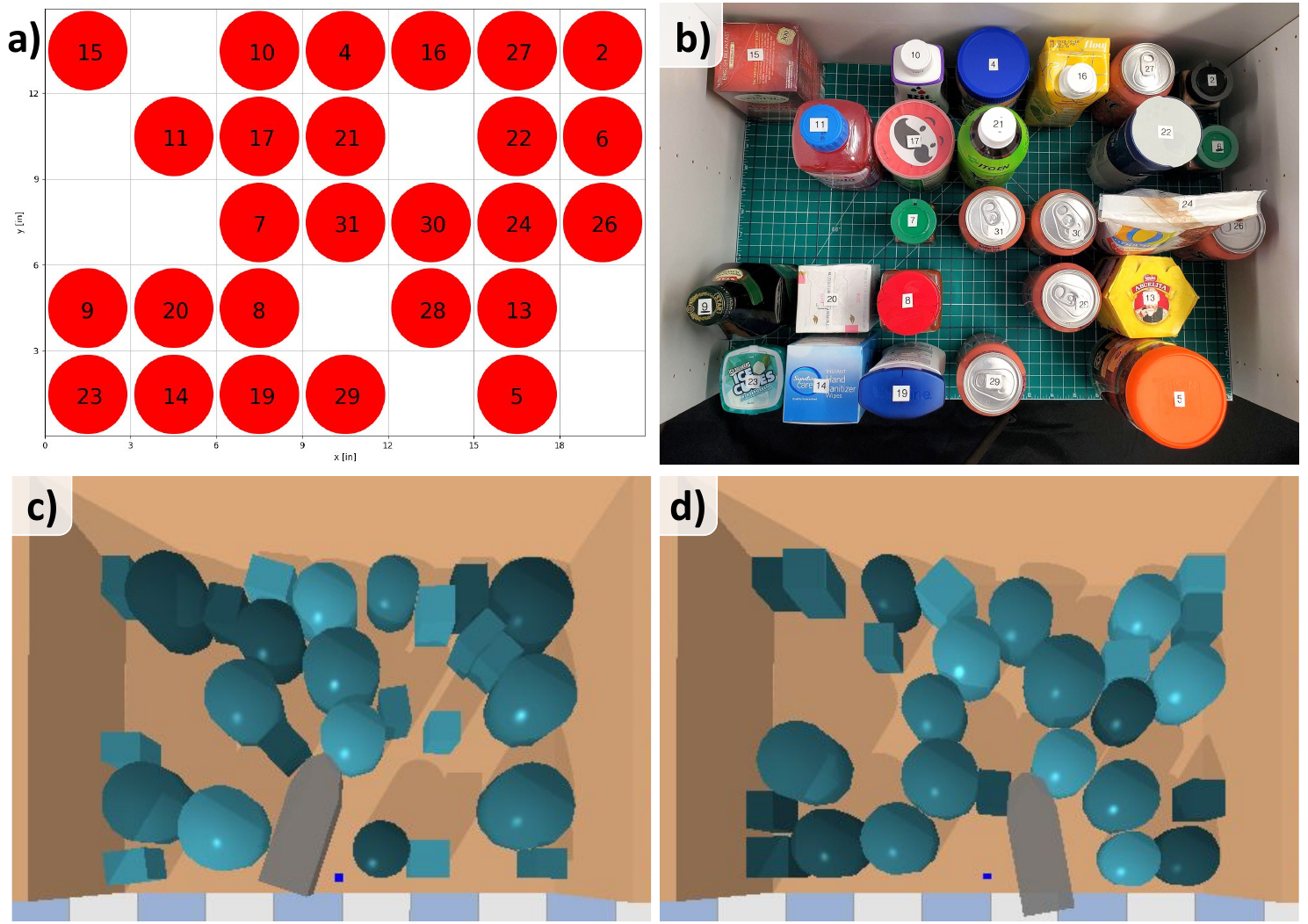}
	\caption{a) Pseudo-randomly generated scene layout on a grid with b) corresponding physical setup to enable repeatable comparisons between control strategies. c-d) Example randomized scenes in PyBullet with darker objects corresponding to heavier objects.}
	\label{fig:setup}
	\vspace{-7pt}
\end{figure}

The scene is divided into a 5$\times$7 grid where individual objects are placed. The total number of objects and specific object IDs for each grid location are pseudo-randomly generated to produce a scene as in \cref{fig:setup}a. The start and goal $x$-locations are randomly selected at the front and back of the scene, respectively. 
Note that although \cref{fig:setup}b shows an overhead view of the scene, a robot reaching into a cabinet does not typically have this view. Specific values for the physical setup, including scene depth, $d_{scene}$ and width, $w_{scene}$, can be found in Table \ref{tab:ParamTable}.

\begin{table}[htb]
\centering
\caption{Hardware Setup Values}
\label{tab:ParamTable}
\begin{tabular}{|c|c|cccc|}
\hline
\multirow{2}{*}{Object Mass} & min & \multicolumn{4}{c|}{143\,g} \\ \cline{2-6} 
 & max & \multicolumn{4}{c|}{570\,g} \\ \hline
\multirow{2}{*}{Object Footprint} & min & \multicolumn{4}{c|}{4.3\,cm $\times$ 4.3\,cm} \\ \cline{2-6} 
 & max & \multicolumn{4}{c|}{8.8\,cm $\times$ 8.8\,cm} \\ \hline
$d_{scene}$ & 38\,cm & \multicolumn{1}{l|}{$v_{max}$} & \multicolumn{1}{c|}{0.045\,m/s} & \multicolumn{1}{c|}{$F_{max}$} & 15.0\,N \\ \hline
$w_{scene}$ & 53\,cm & \multicolumn{1}{l|}{$\omega_{max}$} & \multicolumn{1}{c|}{0.1\,rad/s} & \multicolumn{1}{c|}{$M_{max}$} & 4.5\,N$\cdot$m \\ \hline
\end{tabular}
\end{table}

\subsubsection{Robot Control}

The robot arm used for reaching trials is the Flexiv Rizon 4, which is controlled using a custom Robot Operating System (ROS) architecture. 
The trial workflow includes idle, homing, and reaching states, followed by a resetting state that enables the robot to be re-positioned for another trial once it has reached the goal region or timed out. 
At this time, the scene is manually reset and the next control strategy is deployed. Each control strategy is run on the same scene before cycling to a new scene.

\subsection{Simulation}

We reproduce the hardware setup in simulation to evaluate and refine the motion primitives and enable subsequent performance tests on a larger number of scenes than would be practical on hardware. Example scenes can be seen in \cref{fig:setup}c,\,d. In addition to variations in the precise force thresholds and velocity limits, the primary difference between the simulated and hardware environments is the object placement. In simulation, objects are placed at continuously valued, randomized locations rather than discretized locations on a grid. The simulation code can be accessed here: \textcolor{cyan}{\href{https://github.com/danebrouwer/clutter-jamming.git}{https://github.com/danebrouwer/clutter-jamming.git}}.

\subsection{Experiments}

In the following section, we validate the effectiveness of the primitives in isolation, evaluate whether the performance is robust to primitive parameter variations, and compare the performance of open- and closed-loop hybrid strategies.

The nominal primitive parameter values, ranges for the parameter sweep, and triggering thresholds for the hybrid event control strategy can be found in Table \ref{tab:PrimTable}. Nominal values and triggering thresholds were chosen based on initial qualitative performance on several pilot scenes. The success metrics are final distance to the goal region (a circle with radius 0.75\,cm) and completion time. A successful trial is achieved when the end-effector enters the goal region. 

\begin{table}[h]
\centering
\caption{Control Strategy Values }
\label{tab:PrimTable}
\begin{tabular}{|c|c|c|c|}
\hline
\multicolumn{4}{|c|}{Primitives: Nominal Values} \\ \hline
$A_{bur}$ & 0.83 & \multicolumn{1}{c|}{$f_{bur}$} & \multicolumn{1}{c|}{1\,Hz} \\ \hline
\multicolumn{1}{|c|}{$t_{excv}$} & 5\,s & \multicolumn{1}{c|}{$t_{trig}$}  & 5\,s \\ \hline 
\multicolumn{4}{|c|}{Primitives: Parameter Sweep} \\ \hline
$A_{bur}$ & [0.45:0.05:0.90] & \multicolumn{1}{c|}{$f_{bur}$} & \multicolumn{1}{c|}{[0.5:0.125:1.625]\,Hz} \\ \hline
\multicolumn{1}{|c|}{$t_{excv}$} & [1.875:0.625:7.5]\,s & \multicolumn{1}{c|}{$t_{trig}$}  & [1.875:0.625:7.5]\,s \\ \hline
\multicolumn{4}{|c|}{Hybrid Event Values} \\ \hline
\multicolumn{1}{|c|}{$F_{bur}$} & 5.0\,N & 
\multicolumn{1}{c|}{$F_{excv}$} & 10.0\,N \\ \hline
\multicolumn{1}{|c|}{$F_{push, min}$} & 0.5\,N & 
\multicolumn{1}{c|}{$F_{push, max}$} & 7.5\,N \\ \hline
\multicolumn{1}{|c|}{$t_{push}$} & 2.0\,s & 
\multicolumn{1}{c|}{$t_{prog}$} & 3.0\,s \\ \hline

\end{tabular}
\end{table}

\section{Results}

\subsection{Primitives in Isolation -- Open Loop}
\label{sec:ValPrim}

To validate the effectiveness of the proposed primitives, we evaluate the performance of the three control strategies described in \cref{subsec:OStrats} on the same randomized 
scenes---300 simulated and 25 on hardware.

\begin{figure}[h]
\centering
	\includegraphics[width=2.9in]{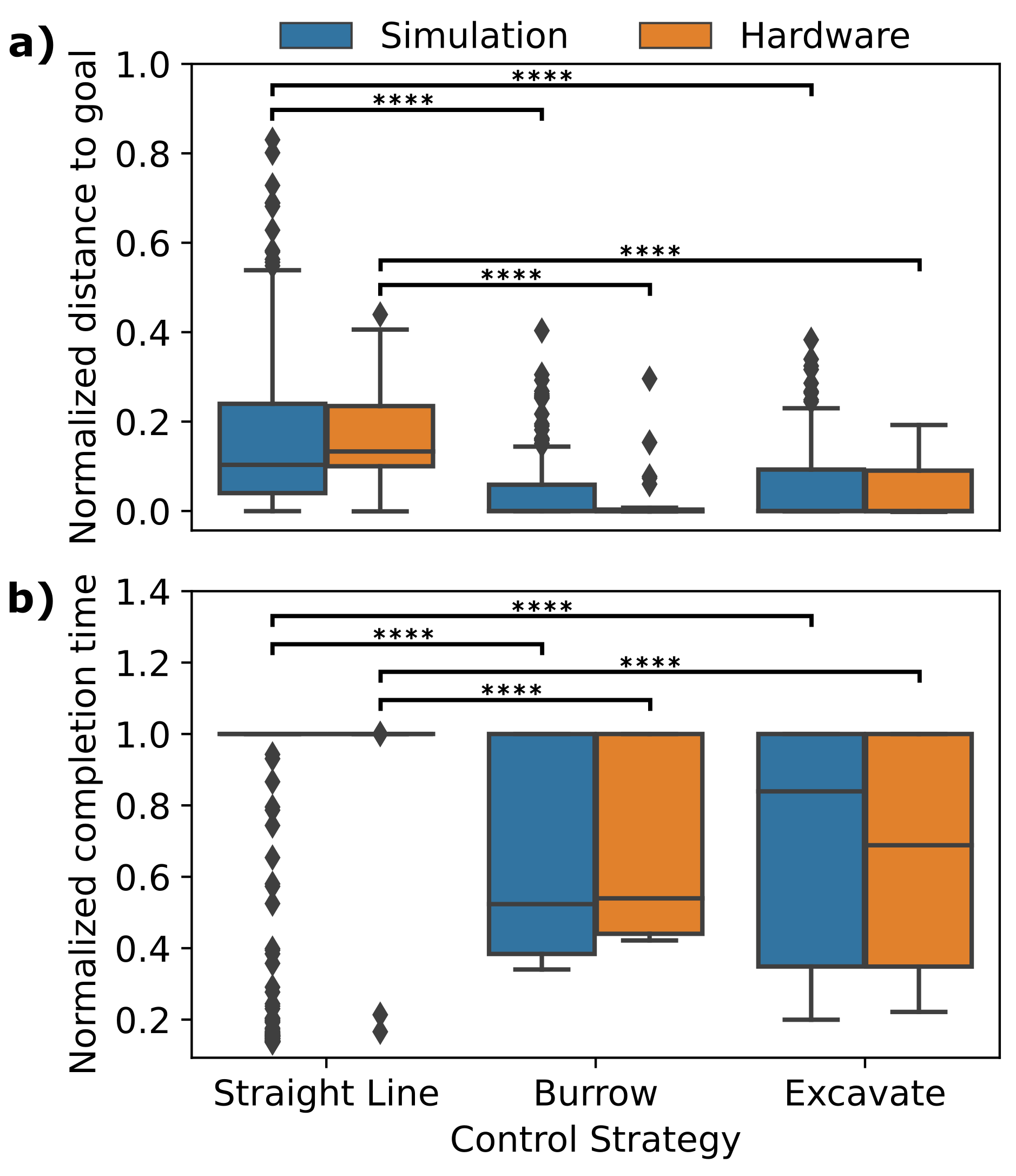}
	\caption{Distributions of a) normalized distance to the goal and b) completion time for the straight line control case and both proposed action primitives on 300 simulated test scenes and 25 physical test scenes. Distance to goal is represented as proportion of scene depth, $d_{scene}$, and completion time as proportion of max duration, $t_{tot}$. **** indicates $p<0.0001$ according to a Wilcoxon signed rank test.
 }
	\label{fig:PrimResults}
	\vspace{-15pt}
\end{figure}

The resulting distributions of final distance to goal and completion time can be seen in \cref{fig:PrimResults}, where normalized distance to goal is computed as $d_{goal}/d_{scene}$ and normalized completion time is $t_{comp}/t_{tot}$.

For the experiments in hardware, straight line, burrow, and excavate success rates are 8\%, 72\%, and 56\%, respectively. For the simulated experiments, the success rates are 22\%, 67\%, and 56\%, respectively, showing a similar trend.

In both simulation and physical trials, the burrow and excavate primitives achieve significantly lower distance to goal and completion time over the straight line case. In these experiments, burrow slightly outperforms excavate, with a lower average distance to goal and completion time as well as higher success rate.

\subsection{Primitive Parameter Variation}

Since the initial hardware and simulated results demonstrate similar trends, we use simulation to provide insights about how performance changes due to parameter variation.
We conduct a parameter sweep to evaluate performance of the burrow and excavate control strategies on 50 scenes for variations in $A_{bur}$, $f_{bur}$, $t_{trig}$, and $t_{excv}$. The values for the parameter sweep can be found in Table \ref{tab:PrimTable}.

\begin{figure}[h!]
\centering
	\includegraphics[width=3.35in]{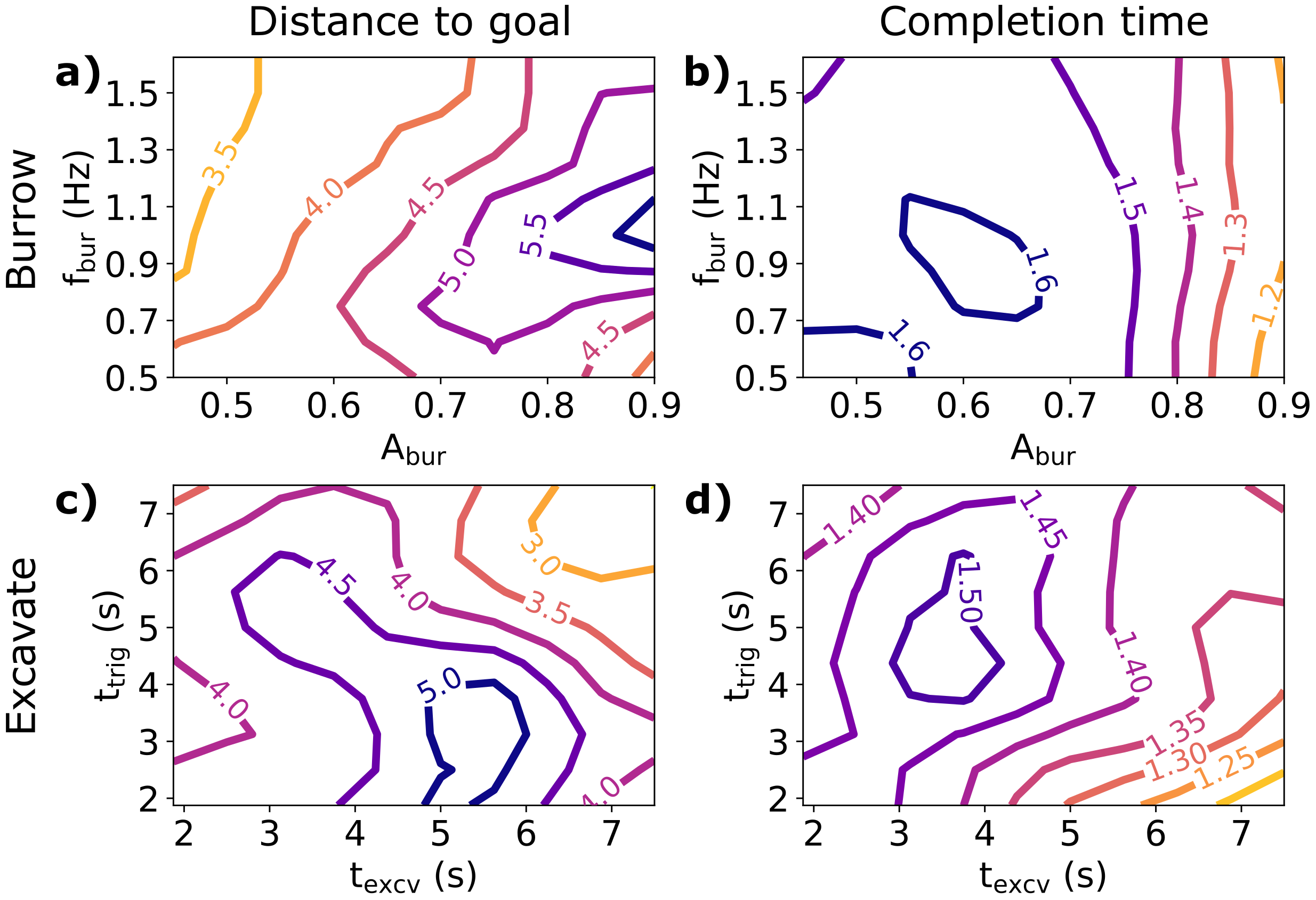}
	\caption {Contour plots with varying primitive parameters showing performance relative to the baseline straight line control strategy for a) burrow distance to goal, b) burrow completion time, c) excavate distance to goal, and d) excavate completion time. Even in the worst cases, the burrow and excavate control strategies substantially outperform the straight line strategy, with $\approx$ 3-fold improvement on distance to goal and $\approx$ 20\% improvement on completion time.  
 }
	\label{fig:PrimParamVar}
	\vspace{-2pt}
\end{figure}

We run the straight line control case on the same 50 scenes in order to compare performance. For each choice of parameters (e.g. $f_{bur} = 1$\,Hz, $A_{bur} = 0.83$), the average performance on these 50 scenes is stored as a ratio of the straight line performance for both metrics. All points for the parameter sweep, now representing a surface, are passed through a Gaussian filter with a standard deviation of $\sigma$ = 1.0. These smoothed surfaces are plotted as shown in \cref{fig:PrimParamVar}, where each contour value is the relative performance above the straight line control strategy.

For the 50 trials, the straight line control strategy has an average normalized distance to goal and completion time of approximately 0.098 and 0.761, respectively. As seen in \cref{fig:PrimParamVar}, burrow and excavate outperform the straight line case in both distance to goal and completion time even when taken at the worst cases, where distance to goal is still improved by $\approx$ 3-fold and completion time is improved by $\approx$ 20\%. The straight line success rate is 38\% on these scenes whereas the worst case success rates for burrow and excavate are 62\% and 68\%, respectively.

The results in \cref{fig:PrimParamVar}a display that if distance to goal is the main objective, low $A_{bur}$ paired with a high $f_{bur}$ should be avoided. \cref{fig:PrimParamVar}b suggests that if completion time is the main objective, a high $A_{bur}$ should be avoided at the cost of distance to goal as noted previously.

For the excavate control strategy, \cref{fig:PrimParamVar}c shows that if distance to goal is the main objective, a high $t_{trig}$ and a high $t_{excv}$ should be avoided. \cref{fig:PrimParamVar}d displays that if completion time is the main objective, a low $t_{trig}$ and a high $t_{excv}$ should be avoided.

\subsection{Hybrid Strategies -- Open-Loop vs. Closed-Loop}

\begin{figure}[h]
\centering
	\includegraphics[width=2.9in]{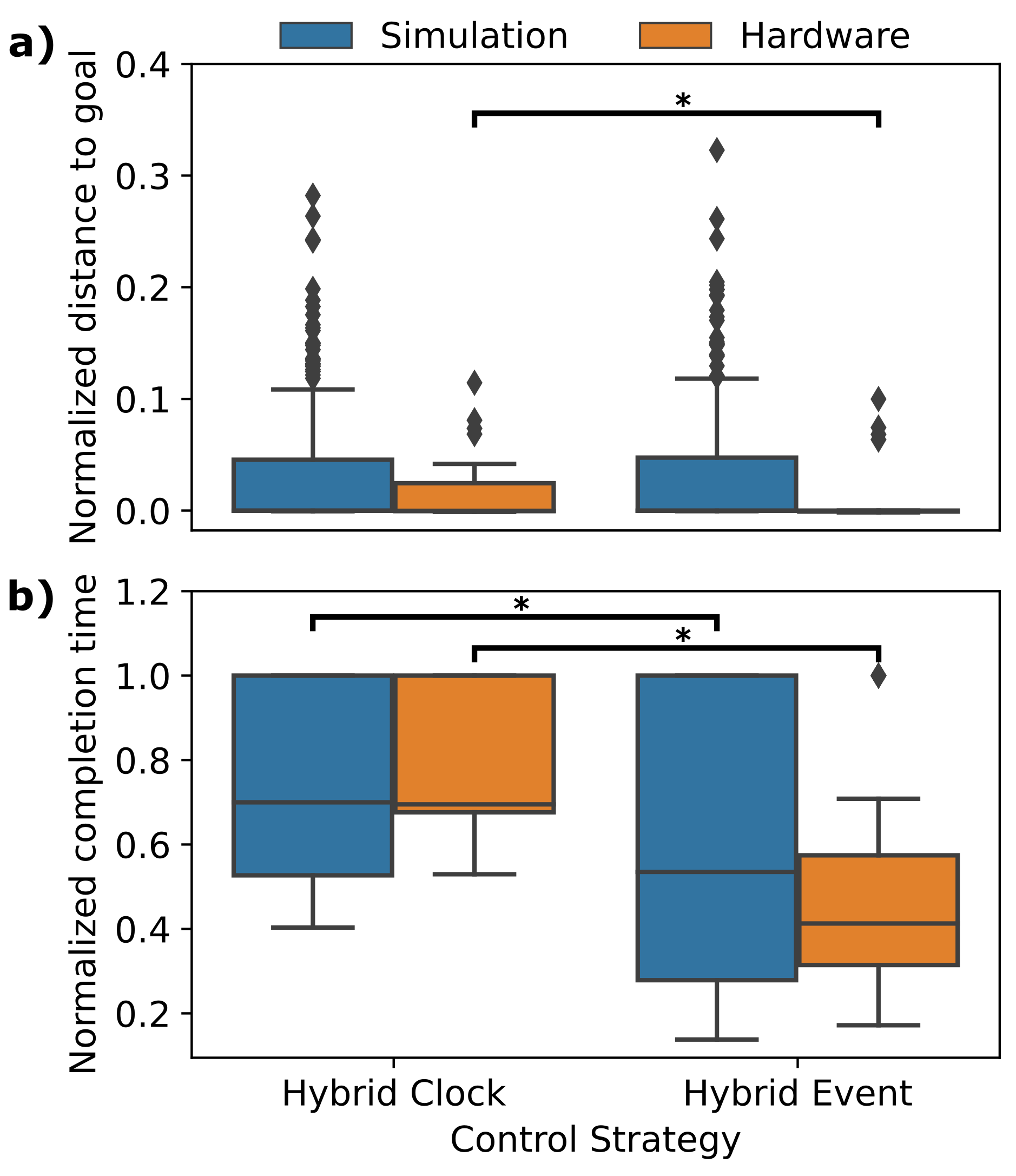}
	\caption{Distributions of a) final distance to the goal and b) completion time for the hybrid clock and hybrid event control strategies on 300 simulated test scenes and 25 physical test scenes. Distance to goal is represented as proportion of scene depth, $d_{scene}$, and completion time as proportion of max duration, $t_{tot}$. * indicates $p<0.05$ according to a Wilcoxon signed rank test. }
	\label{fig:HybridResults}
	\vspace{-15pt}
\end{figure}

The results of evaluating performance on the same 300 simulated and 25 physical scenes as in \cref{sec:ValPrim} for both hybrid strategies can be seen in \cref{fig:HybridResults}, where normalized distance to goal and completion time are computed as before.

Hybrid clock and hybrid event success rates are 72\% and 84\%, respectively, for the hardware experiment. For the simulated experiment, the success rates are 68\% and 71\%, respectively. 
As seen in \cref{fig:HybridResults}a, in tests with hardware, the hybrid event strategy significantly outperforms the hybrid clock strategy in terms of distance to goal. As seen in \cref{fig:HybridResults}b, the completion times for the simulated and hardware tests of the event-driven strategy are both significantly reduced in comparison to their respective clock-driven times. In addition, we observe that during the 25 hardware trials the hybrid clock strategy resulted in 5 items being pushed out of the front of the cabinet whereas for the event-driven strategy this occurred once.

Comparing these results to the previous hardware tests, the hybrid event control strategy significantly outperforms all other control strategies in terms of completion time; it also significantly outperforms the straight line, excavate, and hybrid clock control strategies in terms of the distance to goal. Of the 25 hardware trials, the hybrid event strategy only failed on one scene that any other strategy succeeded on; in contrast, every other strategy failed on at least four scenes that were successful for  that the hybrid event strategy.

\section{Discussion}

Although many researchers have acknowledged the ``sim to real'' gap in tasks that involve contacts \cite{liang2020learning,pmlr-v164-church22a}, the simulated results here---despite using a simplified rigid-body model of contacts in PyBullet---generally matched the trends seen in hardware. As such, they were sufficient to alter strategies and predict which would work best. 
This result is consistent with other findings which show that abstracting motion into primitives and sensory information into features can aid in managing uncertainty \cite{morales2007experiment, hogan2020reactive,thomasson2022going}.
At each instant, the important question to determine how to alter motion was the approximate magnitude and location of the peak contact force on the end-effector. Additional details of the interaction forces were less important. We note also that a single force/torque sensor may not be fitting for this task since it would not report multiple contact locations.

Nonetheless, the results in hardware for the event-based hybrid strategy were better than in simulation. Upon investigation, a likely reason is that the actual contact forces, as measured by the soft triaxial contact sensors, had less noise than the simulated contacts and hence worked better as triggers for the excavate primitive. 
A second point of interest is that while the open-loop strategies---especially the burrow strategy---work well most of the time, they experienced more failures than the event-driven strategy, especially in hardware. Furthermore, the primary advantage provided by tactile information is in swift responses to contact phenomena which enabled faster reaching than all other strategies investigated.

\section{Conclusion and Future Work}

This work furthers a paradigm that embraces, rather than avoids, full-body contact with the environment. This mindset is advantageous for reaching target locations swiftly in densely cluttered scenes since it is difficult to map and plan minimally disruptive paths in constrained, occluded spaces. In this paper, we present readily adoptable, generalizable approaches to enable interaction with these unstructured environments.

We propose the use of two action primitives---burrow and excavate---for reaching toward target locations in dense clutter. We demonstrate that the primitives achieve
a substantial improvement in mitigating jamming across a range of parameters. 
In these actions, soft tactile sensors are advantageous to reduce impulsive contact forces and detect contact locations and forces on the end-effector. These data inform an event-driven hybrid strategy that combines the primitives with a baseline straight motion. In comparison, an open-loop combination of the strategies is both slower and more prone to unproductively disturbing the arrangement of objects.

The findings in this study reveal several possible areas of expansion and improvement. Investigating performance as the scene and object dimensions change will inform whether these approaches are useful in a wide range of scenarios. The proposed primitives can be tools for future researchers using reinforcement learning to learn higher level plans since using motion primitives as discrete action spaces has been shown to improve performance and learning rate \cite{zhang2022learning}.
Optimizing these strategies may improve performance and elicit multi-objective trade-offs that enable tasks to be completed according to the requirements of the application. 

Expansion to multiple sensorized links, 3D clutter, and hardware to acquire objects remains to be demonstrated. Further investigation of which tactile features most benefit the mitigation of jamming and how tactile data compares to other sensory modalities, including vision, may inform system design for robotic interaction with clutter.
Finally, including fixed obstacles may increase the need for tactile-informed planning rather than purely reactive strategies and storing a time history of contacts to characterize the scene is likely essential toward this end.



\section*{ACKNOWLEDGMENT}

Toyota Research Institute provided funds to support this work. D. Brouwer was supported by a Stanford Graduate Fellowship. The authors thank Rick Cory, Hongkai Dai, Rachel Thomasson, Amar Hajj-Ahmad, and Kenneth Hoffmann for insightful discussions.


\bibliographystyle{IEEEtran}
\bibliography{References}

\clearpage

\end{document}